\documentclass[journal]{IEEEtran}

\usepackage{url}
\usepackage{graphicx}
\usepackage{multicol}
\usepackage{color}
\usepackage{amsmath}
\usepackage{multirow}
\usepackage{hyperref}
\usepackage{xcolor}
\usepackage{soul}
\usepackage[switch, right]{lineno}

\newif\ifhighlight
\highlightfalse 
\sethlcolor{yellow} 

\newcommand{\myhl}[1]{%
  \ifhighlight
    \hl{#1}%
  \else
    #1%
  \fi
}

\newcommand\myparagraph[1]{ \vspace{4pt} \noindent \textbf{#1.}}
\newcommand{\paragraphit}[1]{\textit{#1.}}

\begin{document}

\title{Granularity at Scale: Estimating Neighborhood \myhl{Socioeconomic Indicators} from High-Resolution Orthographic Imagery and Hybrid Learning}

\author{Ethan~Brewer,
        Giovani~Valdrighi,
        Parikshit~Solunke,
        Joao~Rulff,
        Yurii~Piadyk,
        Zhonghui~Lv,
        Jorge~Poco,~\IEEEmembership{Member,~IEEE,}
        and~Claudio~Silva,~\IEEEmembership{Fellow,~IEEE}%
\thanks{E. Brewer is with Spectral Sciences, Inc., Burlington, MA, USA. E-mail: ebrewer@spectral.com}%
\thanks{G. Valdrighi and J. Poco are with Fundação Getúlio Vargas, Rio de Janeiro, RJ, Brazil. E-mails: \{giovani.valdrighi, jorge.poco\}@fgv.br.}%
\thanks{P. Solunke, J. Rulff, Y. Piadyk, and C. Silva are with New York University, New York, NY, USA. E-mails: \{parikshit.s, jlrulff, ypiadyk, csilva\}@nyu.edu.}%
\thanks{Z. Lv is with William \& Mary, Williamsburg, VA, USA. E-mail: zlv@wm.edu.}%
}


\maketitle

\begin{abstract}
Many areas of the world are without basic information on the \myhl{socioeconomic} well-being of the residing population due to limitations in existing data collection methods. Overhead images obtained remotely, such as from satellite or aircraft, can help serve as windows into the state of life on the ground and help ``fill in the gaps'' where community information is sparse, with estimates at smaller geographic scales requiring higher resolution sensors. Concurrent with improved sensor resolutions, recent advancements in machine learning and computer vision have made it possible to quickly extract features from and detect patterns in image data, in the process correlating these features with other information. In this work, we explore how well two approaches---a supervised convolutional neural network and semi-supervised clustering based on bag-of-visual-words---estimate population density, median household income, and educational attainment of individual neighborhoods from publicly available high-resolution imagery of cities throughout the United States. Results and analyses indicate that features extracted from the imagery can accurately estimate the density (R$^2$ up to 0.81) of neighborhoods, with the supervised approach able to explain about half the variation in a population's income and education. In addition to the presented approaches serving as a basis for further geographic generalization, the novel semi-supervised approach provides a foundation for future work seeking to estimate fine-scale information from \myhl{aerial} imagery without the need for label data.
\end{abstract}


\IEEEpeerreviewmaketitle

\section{Introduction}
\label{sec:intro}
Censuses and other surveys administered to collect socioeconomic data are expensive and time-consuming \cite{castro_brazil_2023}. For this reason, there is often an undesirably long gap between surveys in \myhl{developing} countries, hindering the appropriate formulation of public policies. The ability to measure socioeconomic metrics is essential for evaluating progress toward targets (such as the United Nations' Sustainable Development Goals \cite{un_susdev}), promoting accountability, enabling evidence-based decision-making, and providing a basis for informed actions and interventions to improve human well-being \cite{psaki_ses_2014,wilson_susdev_2007}. Measurements help determine areas and populations that require the most attention and resources \cite{sapena_metrics_2020}. By quantifying development indicators such as urbanization, education levels, healthcare access, and income, policymakers and development practitioners can identify the most vulnerable and disadvantaged groups and design targeted interventions to address their specific needs and optimize resource allocation \cite{roland_econdev_2017,sapena_metrics_2020,worldbank_conflict_2020}.
\par
Such metrics are traditionally measured through national accounts data, household surveys, and administrative records such as tax filings \cite{castro_brazil_2023}. \myhl{Even in regions where such data are collected, they are expensive \hbox{\cite{edmonstonbarry}}. Furthermore, at the neighborhood level, socioeconomic conditions can undergo drastic changes in very short periods of time \hbox{\cite{SchnakeMahl2020}}. Hence, it is imperative to explore faster and more cost-effective techniques for estimating vital socioeconomic outcomes between neighborhoods.} Since the 1990s, analysis of nighttime light intensity from remote sensing technologies, such as from sensors onboard satellites and aircraft, has effectively contributed to approximating development and socioeconomic metrics at large scales \cite{elvidge_dmspdev_1997,elvidge_dmspdev_2009}. Beginning around 2016, analysis of remotely sensed data (often high-resolution\footnote{We define high-resolution to be each pixel dimension representing a geographic length of $\leq$3 meters.} daytime imagery) with machine learning methods, particularly neural networks, has rapidly grown in popularity and enabled finer scale approximations. These techniques have broadly illustrated that analyzing \myhl{aerial} imagery with machine learning is an effective strategy to remotely monitor the natural and built environment and to estimate and track development and \myhl{socioeconomic statistics} \cite{burke_susdev_2021}. \myhl{Technical challenges arise in this line of research as aerial images can cover vast areas and, at higher resolutions, be too large to input directly into computer vision networks. This complexity is further compounded by the often irregular shapes of neighborhoods, and the aggregation of survey statistics at different levels across various geographical areas.}
\par
\myparagraph{Our objective} In this paper, we investigate the potential of machine learning models trained on high-resolution \myhl{aerial} imagery to estimate the following metrics at the U.S. Census block group level (approximately the size of a neighborhood):
\begin{itemize}
    \item Population density
    \item Median Household Income (MHI)
    \item Educational attainment (\% of the population with at least a bachelor's degree)
\end{itemize}
\myhl{This paper is a feasibility study in how well aerial photography and machine learning can detect where and how people live between neighborhoods. UN sustainable development goals that may benefit specifically from the measurement of neighborhood density, income, and education conditions include goals 1 (poverty reduction), 4 (well-being improvement), and 10 (within-country inequality reduction) \hbox{\cite{un_susdev}}}.
\par
We carry out our investigation by training and testing models on 94 of the 100 largest U.S. cities by gross domestic product (GDP). By automatically extracting spatial features in urban settings, variations in city infrastructure, such as roads, parks, and buildings, can be quantified and related to the census variables. Two methodologies are employed---one driven by a supervised convolutional neural network (CNN) and the other by a semi-supervised framework utilizing bag-of-visual-words (BoVW) to generate simplified but interpretable representations of census blocks.
\par
The most notable \textbf{contributions} of this paper are:
\begin{enumerate}
    \item Demonstration of the ability of contemporary \myhl{aerial} imagery to resolve features related to \myhl{socioeconomic variables} at the scale of a neighborhood;
    \item Finding that supervised learning and \myhl{semi-}supervised clustering of image patches can respectively explain 81\% and 61\% of the variation in neighborhood population density.
\end{enumerate}
\par
\myparagraph{Paper structure} The remainder of this paper is organized as follows. Section \ref{sec:relwork} provides an overview of the existing literature on estimating poverty, population, and other socioeconomic indicators from \myhl{aerial} imagery with machine learning. In Section \ref{sec:data}, we detail how the image and annotation data are acquired, fused, and processed. We describe our methods in Section \ref{sec:methods}, including supervised and semi-supervised learning approaches. Next, we present results in Section~\ref{sec:results}, and analyze and discuss the limitations of our work in Section \ref{sec:discussion}. Finally, concluding remarks are offered in Section \ref{sec:conclusion}.

\section{Related Work} 
\label{sec:relwork}

Throughout much of the world, there is a lack, or a complete absence, of data on the social and economic well-being of people due to conflict, natural disasters, pandemics, and the effort, expense, and time periods between surveys \cite{jean_poverty_2016,castro_brazil_2023}. In recent years, remotely sensed images in combination with machine learning have helped fill in these critical information gaps. \myhl{Use of such techniques has extended into the estimation of population \hbox{\cite{hu_popindia_2019}}, wealth \hbox{\cite{yeh_africa_2020}}, poverty \hbox{\cite{jean_poverty_2016}}, conflict \hbox{\cite{goodman_conflict_2021}}, migration \hbox{\cite{runfola_migration_2022}}, education \hbox{\cite{runfola_testscores_2022}}, land use \hbox{\cite{stoian_landuse_2019}}, and infrastructure \hbox{\cite{brewer_roads_2021,vanetten_spacenet_2021}}, among other applications \hbox{\cite{helber_eurosat_2019,lv_pyshore_2023}}. In this section we focus on studies related to poverty, population density, education, and related metrics. Income and wealth are correlated with self-reported happiness and well-being \hbox{\cite{asadullah_bangladesh_2012,helliwell_happy_2022}}. Post-secondary educational attainment is associated with higher levels of income \hbox{\cite{nces_edattain_2023,zhan_singlemom_2004}}, satisfaction with life \hbox{\cite{jongbloed_euroed_2018}}, and lifelong well-being \hbox{\cite{zhan_retirement_2002}}. Educational attainment for women of reproductive age is linked to reduced child and maternal mortality, lower fertility, and improved reproductive health \hbox{\cite{graetz_africaed_2018}}. Existing studies show a mixed correlation between population density and quality of life \hbox{\cite{greenberg_density_2023,cramer_density_2004,mouratidis_oslo_2019}}. Findings within the city of Oslo, Norway suggest that, compared to residents of lower-density neighborhoods, residents in higher-density neighborhoods have higher levels of personal relationship satisfaction and perceived physical health, similar levels of leisure satisfaction, but lower levels of emotional response to neighborhood and higher levels of anxiety driven primarily by noise and safety concerns \hbox{\cite{mouratidis_oslo_2019}}.}
\par
\myparagraph{Poverty} Gaining momentum in the mid-2010s, several studies have focused on estimating poverty. In \cite{xie_poverty_2016}, a fully convolutional neural network was trained to predict nighttime light intensity from daytime imagery, simultaneously learning features that are useful for poverty prediction at 1 km resolution in Uganda. The model identified different terrains and man-made structures, including roads, buildings, and farmlands, without supervision beyond nighttime lights. Their results approached the predictive performance of survey data collected in the field. \cite{jean_poverty_2016} showed that a CNN trained on Google Maps daytime imagery and existing survey data can identify image features that can explain 37-75\% of the variation in local-level economic outcomes such as wealth and consumption across countries in Africa. In \cite{zhao_multisourcepov_2019}, household wealth in Bangladesh was estimated at 10 km resolution with random forest regression from multiple sources such as nighttime lights, daytime imagery, and land cover maps. \cite{yeh_africa_2020} showed multispectral 30-m Landsat imagery can help estimate African village wealth in countries where the model was not trained with errors comparable to existing ground data. \myhl{Other related work to identify poverty via remote sensing and machine learning include \hbox{\cite{roy_slums_2020}} which validated a Slum Severity Index using Grey Level Co-occurrence Matrix (GLCM) features extracted from high-resolution satellite images of Mexico, \hbox{\cite{luo_povertymaps_2022}} which used high-res imagery and geospatial covariates to characterize degrees of intra-urban deprivation in Nairobi, Kenya (R$^2=0.65$), and \hbox{\cite{arribasbel_deprivation_2017}} in which deprivation in Liverpool, UK was measured by extracting features from Google Earth images (R$^2= 0.54$). Additional related poverty studies include} \cite{sandborn_population_2016,pandey_multitaskpov_2018,li_housing_2021}, and \cite{engstrom_wellbeing_2022}.
\par
\myparagraph{Population} Commonly used techniques for small-area population estimation typically redistribute population ``top-down'' from higher to lower administrative units using areal weighting interpolation or dasymetric mapping techniques \cite{engstrom_density_2020}. Open-source population products that use this approach include Landscan, Meta’s High Resolution Settlement Layer (HSRL), Gridded Population of the World (GPW), WorldPop, and Global Human Settlement Layer (GHSL). Existing studies have focused on redistributing population counts using a random forest-based weighting scheme in Cambodia, Vietnam, and Kenya \cite{stevens_popmap_2015}, redistributing population density in Peru using satellite imagery-based covariates employing regression and tree-based methods \cite{anderson_popden_2014}, and downscaling population counts using one billion mobile phone call records from Portugal and France \cite{deville_mobilepop_2014}. Most population density studies do not validate the accuracy of their estimates against a census \cite{engstrom_density_2020}. Other studies have used coarse nighttime lights for large administrative areas \cite{sutton_censusheaven_2001}, 3D city models \cite{biljecki_popnether_2016}, or focused on a subset of the population (such as children under 5 years of age) \cite{alegana_popmap_2015}. Moving further, \cite{engstrom_density_2020} estimated local population density for in-between census years in Bangladesh by combining household surveys with geospatial data, including an assortment of satellite imagery-based indicators. The data were analyzed with Poisson regression models, with out-of-sample results approximating the density of sub-districts (larger than a village) with an R$^2$ of up to 0.83.
\par
\myparagraph{Additional Metrics} Other closely related work includes \cite{bai_siameseincome_2020} in which a Siamese-like Convolutional Neural Network, integrating ridge regression and Gaussian process regression, was developed for the estimation of income for districts and zip codes in New York City. Their model makes use of a pairwise comparison of location-based house price information, daytime satellite images, street views, and spatial location information, achieving an R$^2$ of 0.72 at the census tract level. \cite{castro_brazil_2023} used daytime and nighttime satellite imagery and transfer learning to estimate average income, GDP per capita, and a water index at the city level in two Brazilian states, explaining up to 64\% of the variation in the target variables. \myhl{In \hbox{\cite{gebru_streetview_2017}}, the authors estimate American Community Survey socioeconomic variables such as income,  race,  education,  and  voting  patterns in 200 US cities at the zip code and precinct level solely through 50 million images of street scenes from Google Street View and computer vision detection of the make, model, and year of all motor vehicles present in the images.} Finally, \cite{graetz_africaed_2018} explored educational inequalities across Africa by estimating years of schooling across a 5x5 kilometer grid based on geocoded survey data, generating estimates of average educational attainment by age and sex.
\par
\myparagraph{Bag-of-visual-words} Widely used in natural language processing, bag-of-words is a numerical representation of text by counting individual words~\cite{joachims1998text}. Despite being a simple formulation, this methodology has shown positive results in diverse language tasks. Inspired by it, computer vision studies have proposed an adaption called bag-of-features or bag-of-visual-words that have shown positive results in natural scene classification~\cite{coates_analysis_2011, bosch_scene_2006}. By creating a set of low-level visual ``features'' that describe the images, the frequency of the visual features in each image can be used for predictive tasks. The method has also shown positive results~\cite{verikas_high_2017, kalinicheva_satautoencoder_2020, metzler_phenotyping_2023} in the domain of remote sensing imagery. Despite this existing work, these techniques have not been tested to estimate high-resolution census variables.
\par
Most existing remote sensing studies analyze variables for entire nations, states, or cities, potentially obscuring neighborhood-level prosperity and inequality patterns. Our study pushes beyond the limitations of existing work by investigating population density, income, and education at an unprecedentedly precise scale across a country, using free, publicly available data.

\section{Data}
\label{sec:data}

In this section, we detail how the data for the 94 cities are collected and processed.

\myparagraph{Imagery} For the United States, orthographic imagery is retrieved from the National Agriculture Imagery Program (NAIP) \cite{usda_naip}, administered by the U.S. Department of Agriculture. For a given point in the U.S., RGBIR aerial imagery is acquired approximately every 2-3 years at a resolution of 60-centimeter ground sample distance during the agricultural growing season, or “leaf on” conditions. The images are orthorectified, which combines the image characteristics of an aerial photograph with the georeferenced qualities of a map. We utilize the most recent NAIP tiles (2019-2021) for 94 prominent cities across the United States, including the ten largest (by GDP). Our selection process involved filtering cities of the 100 largest metropolitan statistical areas (MSAs) based on 2021 GDP, followed by identifying the largest city by area within each MSA. This selection is used to study a diverse set of cities while considering the most significant ones.

\myparagraph{Annotation data} Census data for the United States are acquired through the American Community Survey (ACS) \cite{uscensus_acs}. Every year, the U.S. Census Bureau contacts approximately 3.5 million households (1 in 40 total households) across the country to participate in the ACS, with a 2021 response rate of 85.3\%. The survey includes various demographic, social, economic, and housing data on residents such as age, race, occupation, income, disability status, housing type (e.g., single-family, multi-unit), languages spoken, and highest degree earned. The resulting data products are aggregated at various levels from country, to state, to county, to census tract, to block group. In this study, data are examined at the finest possible level, block group, to extract the maximum benefit from the resolution of the imagery. Block groups contain an average of approximately 1,500 residents and may henceforth be referred to as neighborhoods.
\par
The following 5-year\footnote{5-year estimates aggregate data from the preceding 60-month period. For example, a 5-year estimate from the 2021 ACS aggregates data from 01Jan2017 to 31Dec2021 \cite{uscensus_5year}.} ACS variables for neighborhoods are downloaded via API for all counties containing the cities (since the ACS aggregates by county but not city) from the year their associated imagery was captured:
\begin{itemize}
    \item Total population, $P_{t}$
    \item Population $>$25 years old, $P_{25}$
    \item Median Household Income, MHI
    \item Four education variables: Population $>$25 years old whose highest degree completed is (1) bachelor's, $P_b$, (2) master's, $P_m$, (3) professional, $P_p$, (4) doctoral, $P_d$
\end{itemize}
\noindent
An educational attainment metric, $E$, representing the percent of the population with at least a bachelor's degree is calculated with
\begin{equation}
    E = \frac{P_b+P_m+P_p+P_d}{P_{25}}\times100
\end{equation}

\noindent
The population density metric, $D$, representing people per square kilometer is calculated with
\begin{equation}
    D = 10^6\frac{P_t}{A}
\end{equation}

\noindent
where $A$ is the geographic area of a neighborhood in square meters found by
\begin{equation}
    A = \frac{\textrm{\# nonzero pixels in image}}{\textrm{Resolution of image}}
\end{equation}

\noindent
See Table \ref{tab:crop_data} for all specifications of the neighborhoods analyzed.

\begin{figure*}
    \centering
    \includegraphics[width=\textwidth]{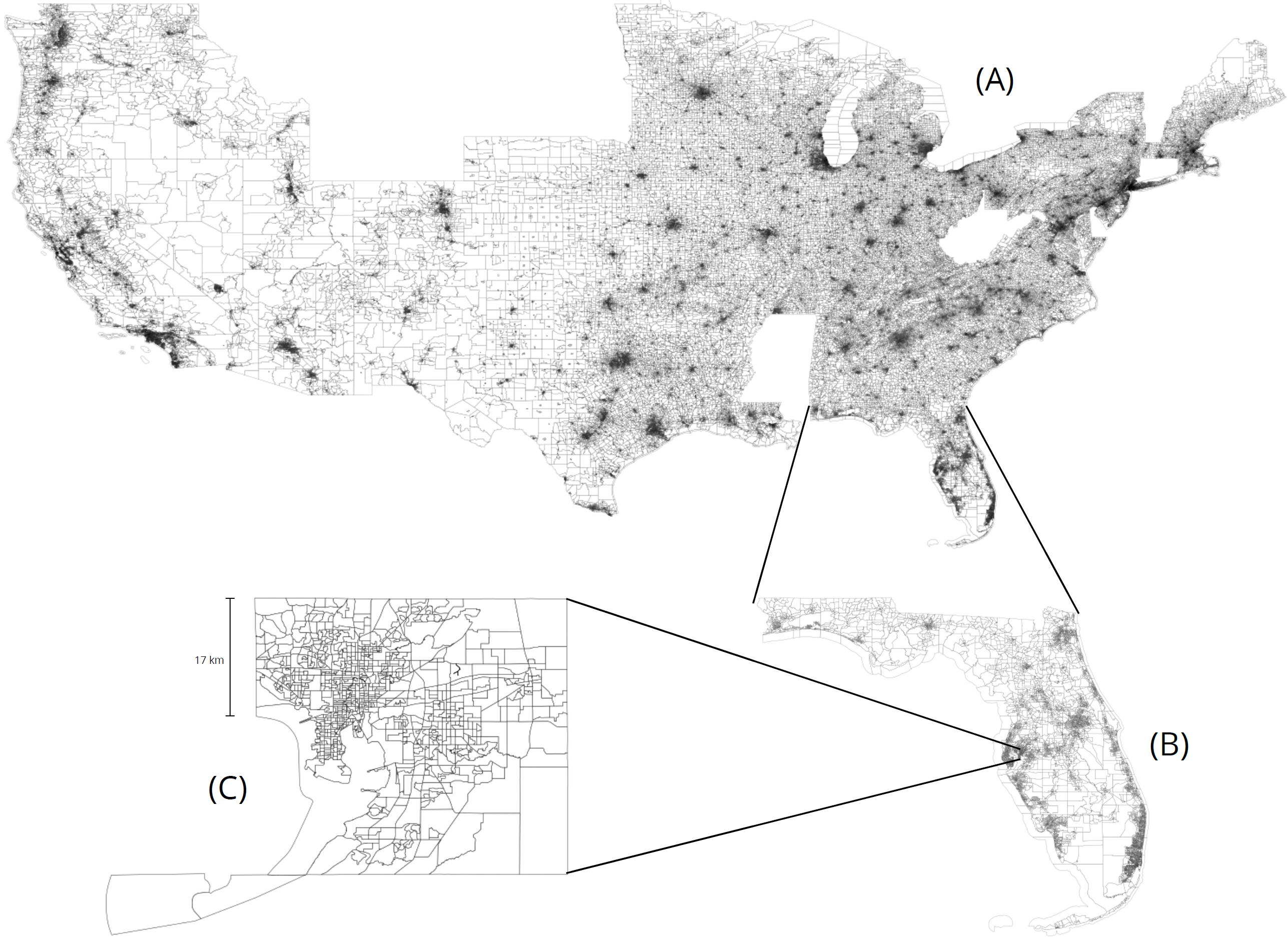}
    \caption{(A) Illustration of the neighborhoods in states containing cities analyzed in this study. (B) Blow up of neighborhoods in the state of Florida. (C) Blow up of neighborhoods in the \myhl{county of Hillsborough, Florida which contains the city of Tampa.}}
    \label{fig:block_groups}
\end{figure*}

\myparagraph{Image-label pairing} To generate geographic boundaries for the imagery, shapefiles of the neighborhoods are downloaded from the U.S. Census Bureau's TIGER archive \cite{uscensus_tiger} (see Fig. \ref{fig:block_groups}). This geographic information (polygons of neighborhoods) is then merged with their corresponding ACS variables. In the process, neighborhoods containing census errors or zero population are dropped.

\begin{figure}[ht!]
    \centering
    \includegraphics[width=252pt]{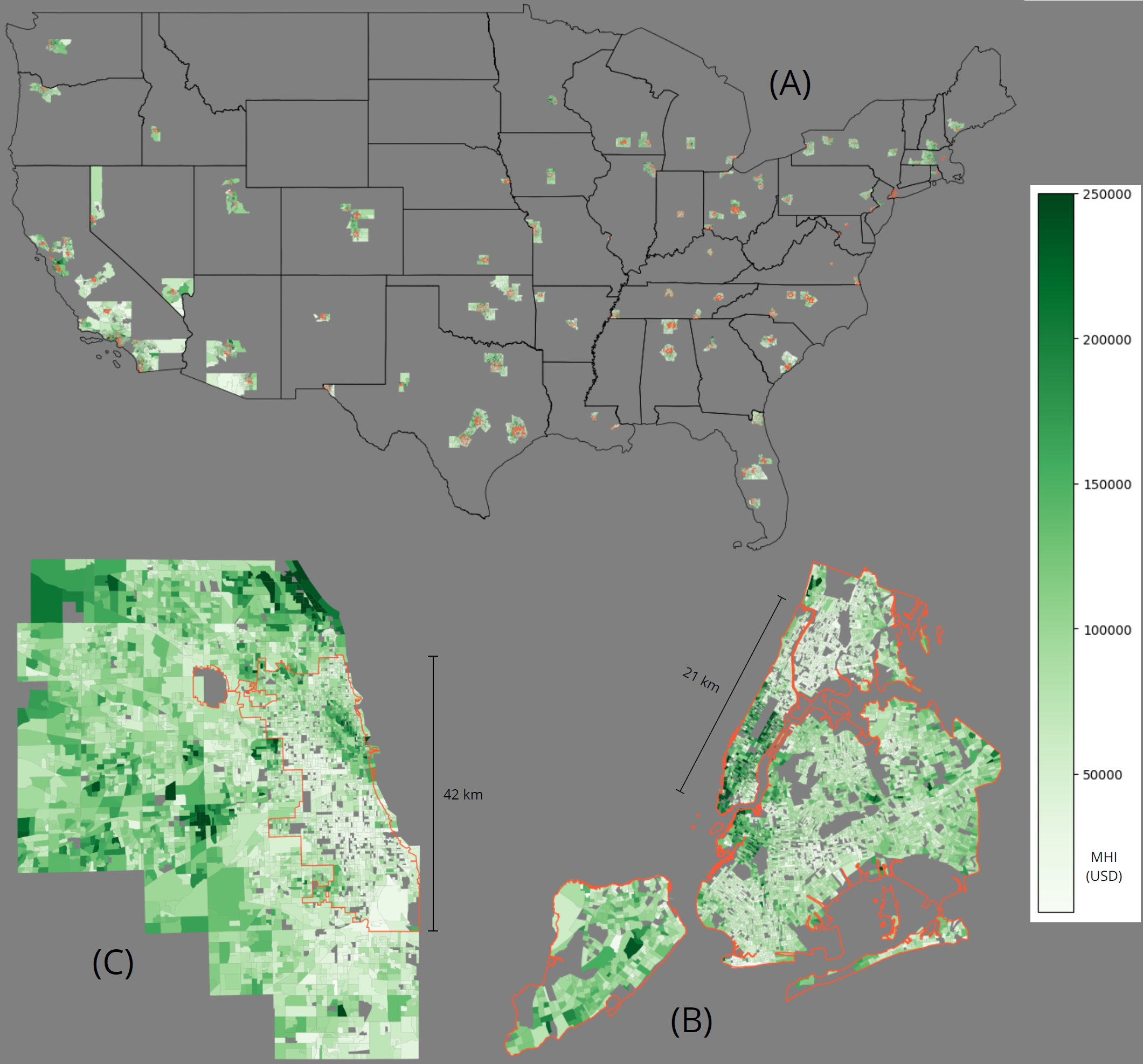}
    \caption{(A) Illustration of median household income (in 2021 USD) of counties containing the 94 cities examined. City boundaries are in orange. (B) Expanded view of New York City in which its boroughs are coterminous with counties. (C) Expanded view of Chicago in which its city limits are within Cook and DuPage counties (mostly Cook).}
    \label{fig:mhi_maps}
\end{figure}

Fig. \ref{fig:mhi_maps}A shows all the cities examined (in orange outlines) overlaid with the ACS median household income by neighborhood, as an illustration. The cities of New York and Chicago are enlarged in Figs. \ref{fig:mhi_maps}B\&C to provide a more detailed view.

Next, the imagery is cropped by neighborhood based on the bounding boxes of the neighborhood polygons. This results in a total of 43,497 images (see Table \ref{tab:crop_data} for other specifications on the neighborhood crops).

\begin{table}[!ht]
    \caption{Neighborhood Specifications}
    \label{tab:crop_data}
    \centering
    \begin{tabular}{|c|c|}
    \hline
    \bf{Specification} & \bf{Value} \\
    \hline
    \# images &  43,497 \\ 
    \hline
    Image width range & 103-17,101 pixels \\ 
    \hline
    Median width & 1,353 pixels \\
    \hline
    $\sigma$ in width & 1,124 pixels \\
    \hline
    Image height range &  137-22,174 pixels \\ 
    \hline
    Median height & 1,350 pixels \\
    \hline
    $\sigma$ in height & 1,141 pixels \\
    \hline
    Density range & 2-47,791 $\frac{\textrm{ppl}}{\textrm{km}^2}$ \\
    \hline
    Median density & 634 $\frac{\textrm{ppl}}{\textrm{km}^2}$ \\
    \hline
    Mean density & 1,457 $\frac{\textrm{ppl}}{\textrm{km}^2}$ \\
    \hline
    $\sigma$ in density & 2,519 $\frac{\textrm{ppl}}{\textrm{km}^2}$ \\
    \hline
    MHI range & 2,499-250,001 USD\\
    \hline
    Median MHI & 63,232 USD \\
    \hline
    Mean MHI & 72,997 USD \\
    \hline
    $\sigma$ in MHI & 42,595 USD \\
    \hline
    Educational attainment range & 0.0-100 \\
    \hline
    Median educational attainment & 31.3 \\
    \hline
    Mean educational attainment & 36.2 \\
    \hline
    $\sigma$ in educational attainment & 24.7 \\
    \hline
    \end{tabular}
\end{table}

\begin{figure}[ht!]
    \centering
    \includegraphics[width=252pt]{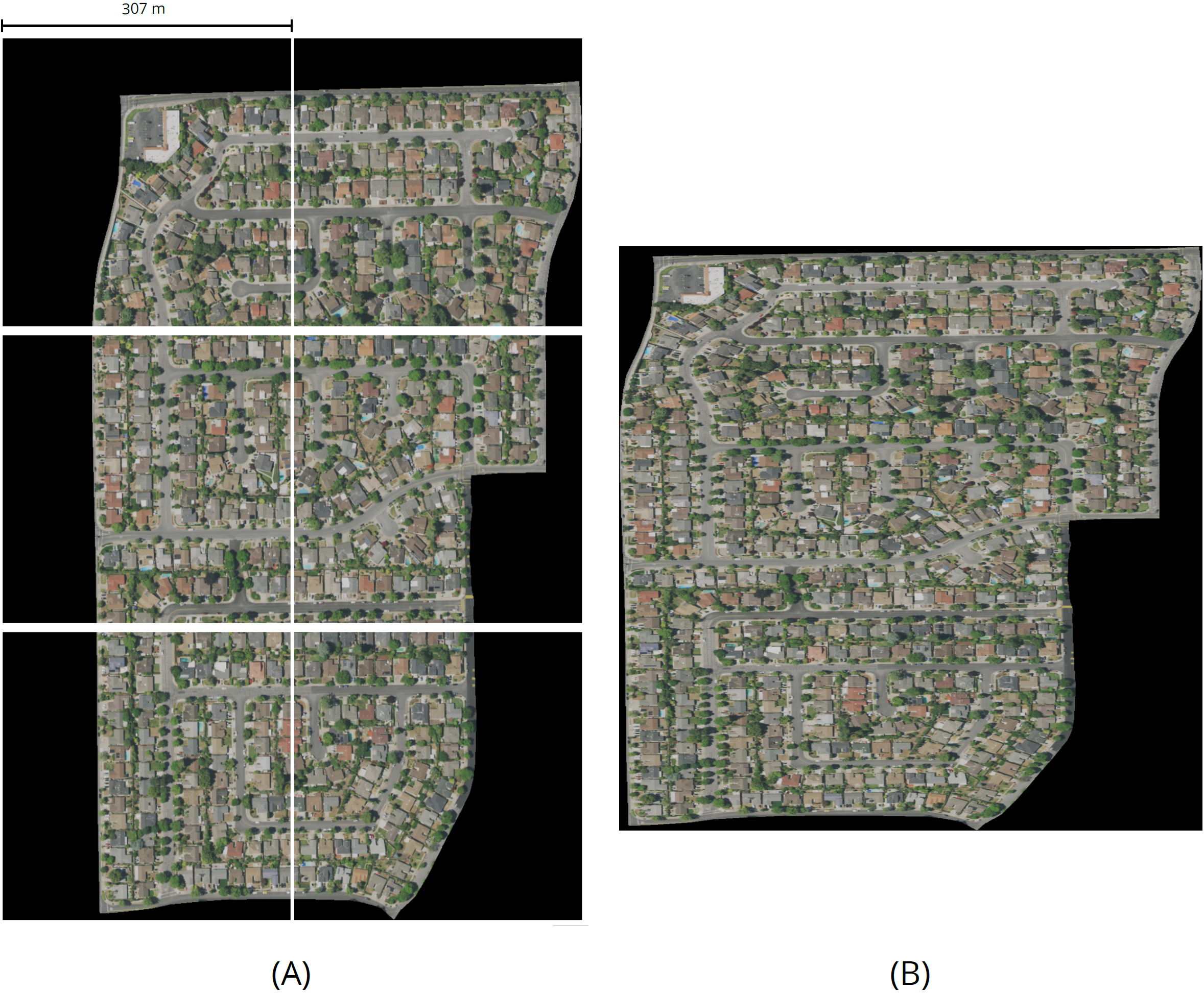}
    \caption{Processing of a typical neighborhood (this one is in San Jose, CA) for the two processing methods for supervised learning. (A) Patching: The image is split into six 512x512 patches. (B) Resizing: The image is resized to 1353x1350 pixels (the median width and height of a neighborhood).}
    \label{fig:supervised_images}
\end{figure}

\myparagraph{Crop processing} The crops are processed for CNN input for the supervised method in two ways, ``patching" and ``resizing" (an example is visualized in Fig. \ref{fig:supervised_images}).

\paragraphit{Patching} With this technique, neighborhoods are split into 512x512 patches, as in Fig. \ref{fig:supervised_images}A. If either of the original dimensions of an image is not a multiple of 512, it is padded by zeros before being split. Only patches composed of $>$50\% nonzero pixels are kept. This results in a total of 339,413 patches. Patching allows an image of a neighborhood to retain its resolution and shape, but results in the CNN treating each patch as a separate image, thus breaking apart the spatial relationship within a neighborhood. 

\paragraphit{Resizing} With this technique, neighborhoods are resized (through bilinear interpolation) to the median size of a neighborhood, i.e., a width of 1353 pixels and a height of 1350 pixels. Resizing allows a neighborhood to be read as a single image by the CNN, but results in upsampling/downsampling for crops that have a dimension(s) less/greater than the median, and shape distortion for crops with a $\frac{\textrm{width}}{\textrm{height}}$ ratio different from $\frac{1353}{1350}$.
\par

\begin{figure}
    \centering
    \includegraphics[width = \linewidth]{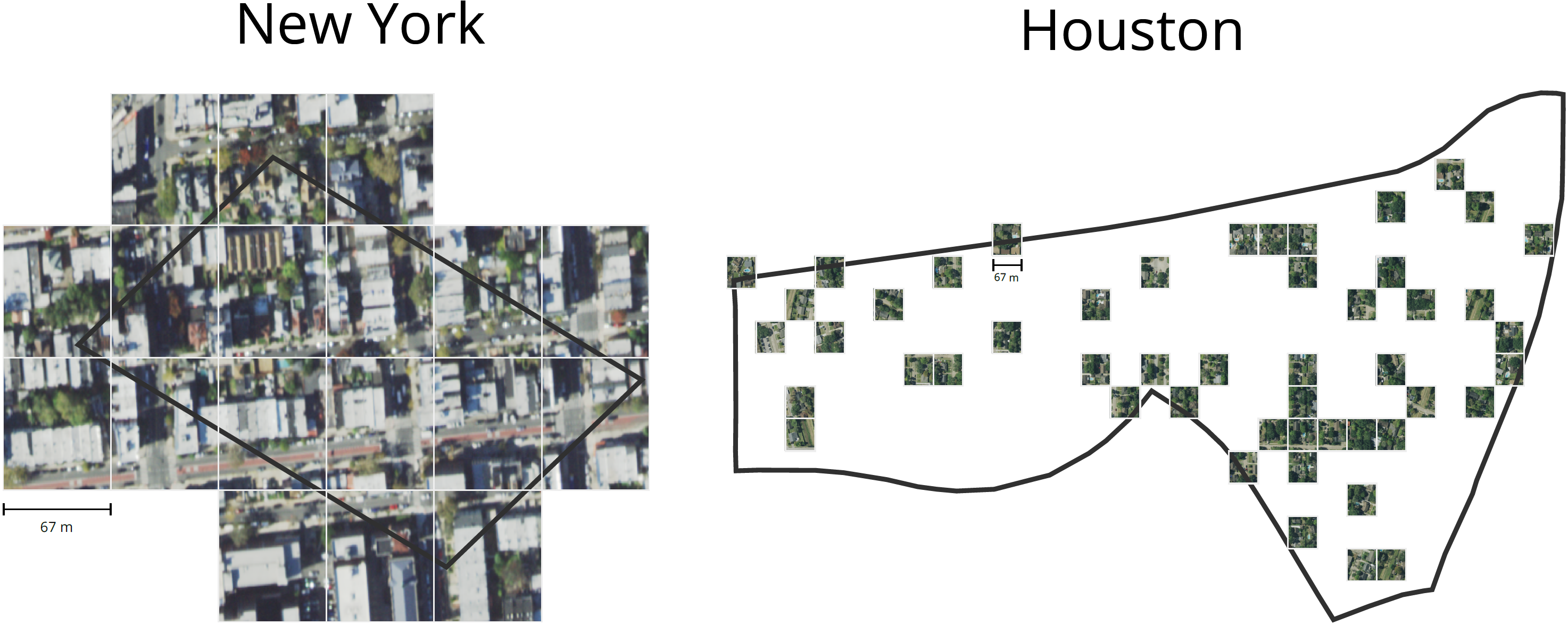}
    \caption{For the semi-supervised approach, example\myhl{s} of 112x112 patches for neighborhoods in different cities. Patch boundaries are denoted with white borders, and census block groups (i.e., neighborhoods) with black borders. In the New York neighborhood, all patches overlapping with the neighborhood are used. For the larger Houston neighborhood, only 50 samples \myhl{are} selected.}
    \label{fig:grid_patches_example}
\end{figure}

\paragraphit{Semi-Supervised} In the semi-supervised methodology, the imagery is cropped into a square grid, each cell measuring 112x112 pixels. Therefore, each neighborhood is composed of a mosaic of patches. This high granularity serves the purpose of separating distinct urban structures within each patch. Due to the NAIP tiles covering areas not examined in this study, only the subset of obtained patches that have an intersection to any neighborhood are used. Additionally, considering that some neighborhoods are very large (i.e., they have low population density), further filtering is implemented to limit the maximum number of patches to 50 per neighborhood to reduce computational costs. The patches are semi-randomly selected at a probability proportional to the percentage of their area contained in the neighborhood. As a result of these filtering measures, around two million 112x112 patches are generated (see Table \ref{tab:tvt_data}). Fig.~\ref{fig:grid_patches_example} depicts two example neighborhoods and their division of patches.
\par
For both supervised and semi-supervised approaches, the resulting data are separated into training, validation, and testing sets for model input in 70-15-15\% splits. Dataset sizes are shown in Table \ref{tab:tvt_data}. 

\section{Methods}
\label{sec:methods}

We now present the formulations of each methodology and their details.

\subsection{Supervised}

\begin{figure*}
    \centering
    \includegraphics[width=\textwidth]{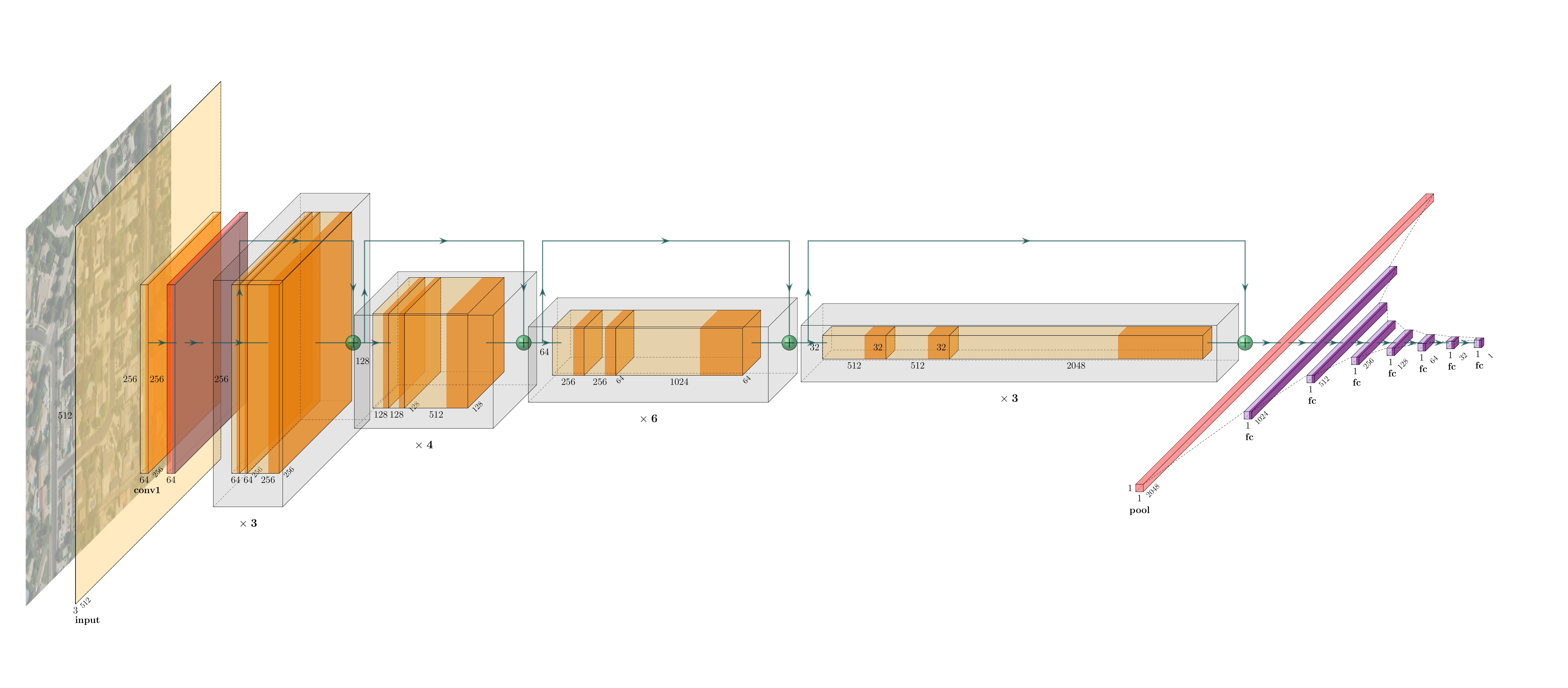}
    \caption{Visual representation of the ResNet50-based architecture used in the supervised approach. 30\% dropout layers are embedded after the first four fully-connected layers.}
    \label{fig:ResNet50_Viz}
\end{figure*}

The overall supervised methodology for both the image patching and resizing techniques is executed in the following manner:

\begin{enumerate}
    \item A ResNet50-based\cite{he_resnet_2016} architecture (shown in Fig. \ref{fig:ResNet50_Viz}) is trained in separate instances on each of the three target variables (density, MHI, education);
    \item The trained networks are evaluated on the independent test set not included in the training process.
\end{enumerate}

The ResNet50-based architecture of Fig. \ref{fig:ResNet50_Viz} has its base model pre-trained on ImageNet.\footnote{A test was conducted with weights pre-trained on classification and segmentation tasks with Sentinel imagery, but it did not perform better than ImageNet weights.} In this study, seven fully-connected layers are added after the base model to gradually scale down the feature space to a single output estimation.

\begin{table}[!ht]
    \centering
    \caption{Dataset Sizes}
    \label{tab:tvt_data}
    \begin{tabular}{|c|c|c|c|}
        \hline
        \multirow{2}{*}{\textbf{Dataset}} & \multicolumn{3}{c|}{\textbf{Method}} \\
        \cline{2-4}
        & Sup. Patching & Sup. Resizing & Semi-sup. \\
        \hline
        Train (\# images) & 237,589 & 30,447 &  1,364,790 \\ 
        \hline
        Validation (\# images) & 50,911 & 6,524 & 294,483 \\
        \hline
        Test (\# images) & 50,913 & 6,526 &  293,322 \\
        \hline
        Total (\# images) & 339,413 & 43,497 & 1,952,595 \\
        \hline
    \end{tabular}
\end{table}

For model training on each metric, updating all weights in all layers for patching and updating only the fully-connected layers for resizing produces optimal results. A batch size of 16, AdamW optimization, and L1 loss (i.e., mean absolute error, MAE) are utilized throughout. For each training epoch, the models are trained on the training set and then accessed with the validation set. Models are trained until there is no improvement in validation accuracy after 5 epochs. Model weight configurations with the lowest validation loss are evaluated on the test set.

\subsection{Semi-supervised}

\begin{figure*}
    \centering
    \includegraphics[width=\textwidth]{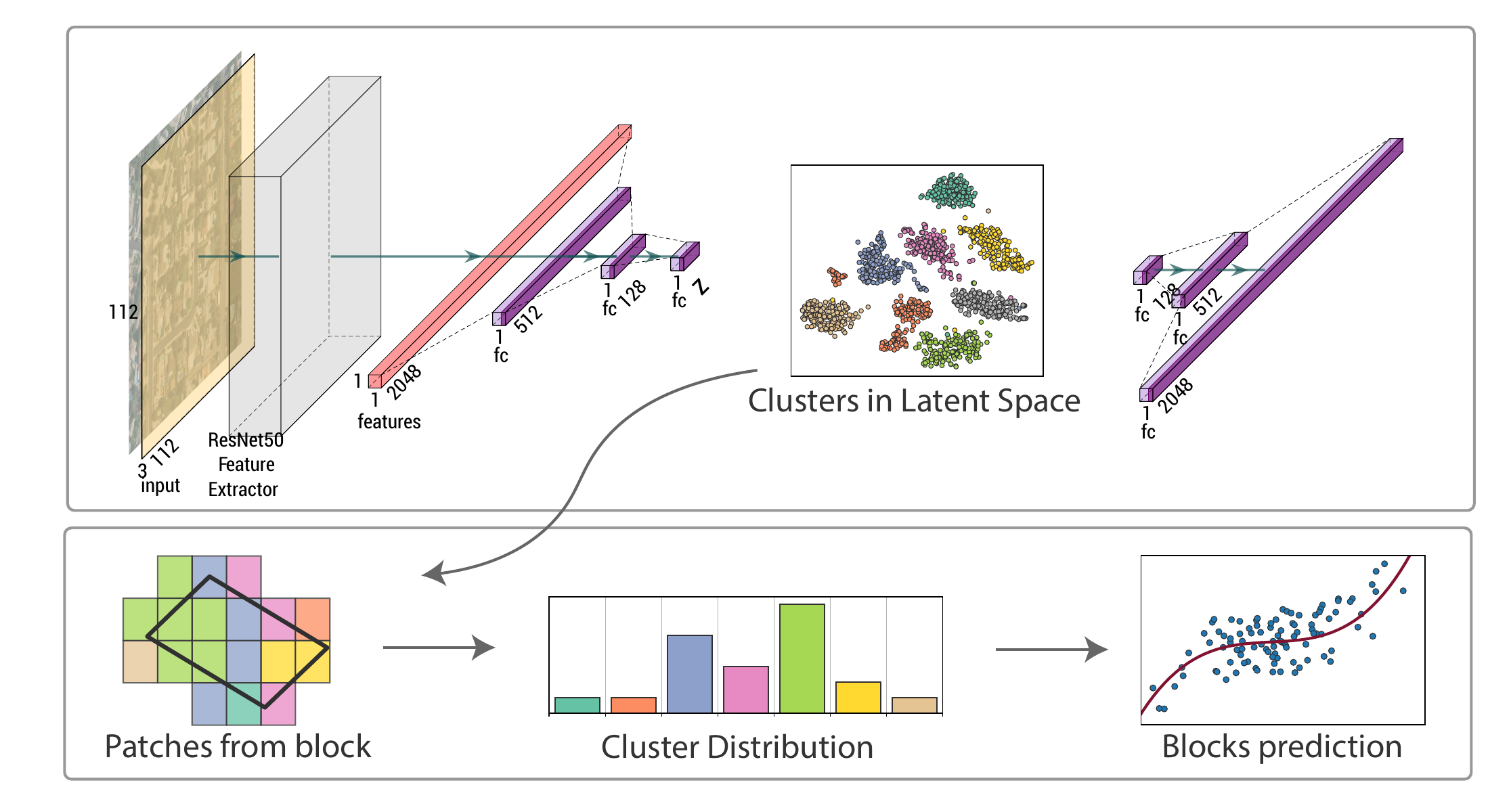}
    \caption{\myhl{Overall} steps of the semi-supervised methodology. First, an unsupervised clustering algorithm is used to cluster small patches of neighborhoods from aerial imagery. The clustering uses ResNet50 as a feature extractor and an autoencoder. The second step is supervised regression on the target variables using the distribution of clusters composed of neighborhood patches.}
    \label{fig:semi_supervised_methodology}
\end{figure*}

As discussed in Section.~\ref{sec:relwork}, we use ideas from bag-of-visual words to produce a rich set of features from the high-resolution imagery that are interpretable and correspond to distinct urban infrastructures. These features are later used to fit a supervised model with the target variables. The methodology (Fig.~\ref{fig:semi_supervised_methodology}) can be separated into two steps: clustering of patches and calculation of the cluster distribution of each neighborhood.
\par
\myparagraph{Clustering of patches} While cities in the U.S. exhibit variations in culture and environment, there are shared characteristics in their urban infrastructure that can be organized into clusters. However, due to the high dimensionality of images and the distances defined between pixel colors, clustering algorithms can present better results when applied to a learned representation of the images. Diverse methodologies have already used the representational power of deep neural networks to improve clustering results~\cite{zhou2022comprehensive}. A simple and effective technique is to run $k$-means in the latent representation learned from an autoencoder. Deep~Embedding~Clustering~(DEC)~\cite{xie_deepembedding_2016} is a more sophisticated technique that trains an autoencoder in two stages: after the first stage optimizes for reconstruction, the embeddings are clustered by $k$-means and the resulting centroids are added as parameters jointly with the encoder. The loss penalizes the distance between the embeddings and their respective centroids in the second stage. 

In our work, we evaluate using both $k$-means in a regular autoencoder and $k$-means in an autoencoder trained with DEC. For both techniques, a ResNet50 pre-trained on ImageNet is used as a feature extractor from patches, generating a 2048-dimension representation for each. The autoencoder is defined as a feed-forward network with the architecture depicted in Fig.~\ref{fig:semi_supervised_methodology}, i.e., four layers in the encoder and decoder. The choice of the latent space dimension ($d_Z$) hyperparameter is crucial to balance the expressive power of the autoencoder and the effectiveness of the $k$-means clustering. A higher dimension allows for lower reconstruction loss, but it may hinder the clustering process since $k$-means relies on the Euclidean distance between embeddings. Therefore, the latent dimension is tested using $\{32, 64\}$. A second necessary parameter is the number of clusters, $k$, which is also selected between $\{50, 100, 200\}$ through experimentation.
\par
\myparagraph{Cluster distribution} In this step, we use the patch clusters to build two different sets of features for the neighborhoods. The first set of features is the frequency of each of the $k$ clusters among the patches. The second set of features calculates the distance in the latent space of the patches to each of the $k$ centroids defined in the latent space. Both methods attempt to describe each neighborhood as a composition of clusters, i.e., a composition of distinguished urban infrastructures. The features then can be used in a regression model. We choose to evaluate with random forest.
\par
As mentioned, the methodology has hyperparameters that need to be selected: the dimension of latent space $d_Z$, the number of clusters $k$, the type of the set of features, and random forest hyperparameters. The autoencoder and random forest training are made using only the training dataset, with hyperparameters selected based on performance on the validation set. We present separate results comparing $k$-means and DEC.

\section{Results}
\label{sec:results}

\subsection{Supervised}

Table \ref{tab:supervised_results} displays supervised results on the test data, in terms of mean absolute error (MAE) and R$^2$, for patching and resizing image processing techniques. In Table \ref{tab:supervised_results}, bold fonts highlight the most accurate MAE and R$^2$ results. Models performed best at measuring density, with models trained on resized neighborhoods able to explain 81\% of the variation in density across the study area. These models were able to estimate density to within 461 $\frac{\textrm{ppl}}{\textrm{km}^2}$, on average (for reference, the density variable has a ground truth standard deviation of 2519 $\frac{\textrm{ppl}}{\textrm{km}^2}$). With resizing, models trained to measure median household income and educational attainment are able to explain about half the variance in the ground truth (R$^2$ of 0.48 and 0.51, respectively). Models trained on patches performed as well as resized images for estimating education level, however, they were about 7-8\% worse (in terms of R$^2$) at estimating density and MHI than their resize-based counterparts. Additionally, it is worth noting that while the R$^2$ score is similar for estimating education level, the MAE is lower using patches compared to resizing. 
f

\begin{table}[!ht]
    \centering
    \caption{Supervised Results}
    \label{tab:supervised_results}
    \renewcommand{\arraystretch}{1.35}
    \begin{tabular}{|c|c|c|c|c|}
        \hline
        \multirow{3}{*}{\textbf{Metric}} & \multicolumn{4}{c|}{\textbf{Method}} \\
        \cline{2-5}
        & \multicolumn{2}{c|}{Patching} & \multicolumn{2}{c|}{Resizing} \\
        \cline{2-5}
        & MAE & R$^2$ & MAE & R$^2$ \\
        \hline
        {Density $\left(\frac{\textrm{ppl}}{\textrm{km}^2}\right)$} & \textbf{142} & 0.73 & 461 & \textbf{0.81} \\
        \hline
        {MHI (USD)} & 23,816 & 0.41 & \textbf{21,579} & \textbf{0.48} \\
        \hline
        {Education (\%)} & \textbf{12.9} & 0.50 & 13.2 & \textbf{0.51} \\
        \hline
    \end{tabular}
\end{table}

\begin{table}[!ht]
    \centering
    \caption{Semi-supervised Results}
    \label{tab:semi-supervised_results}
    \renewcommand{\arraystretch}{1.35}
    \begin{tabular}{|c|c|c|c|c|}
        \hline
        \multirow{3}{*}{\textbf{Metric}} & \multicolumn{4}{c|}{\textbf{Method}} \\
        \cline{2-5}
        & \multicolumn{2}{c|}{$k$-means} & \multicolumn{2}{c|}{DEC} \\
        \cline{2-5}
        & MAE & R$^2$ & MAE & R$^2$ \\
        \hline
        {Density $\left(\frac{\textrm{ppl}}{\textrm{km}^2}\right)$} & \textbf{3,583} & \textbf{0.61} & 3,743 & 0.55 \\
        \hline
        {MHI (USD)} & \textbf{31,366} & \textbf{0.03} & 31,534 & 0.02 \\
        \hline
        {Education (\%)} & \textbf{20.3} & \textbf{0.04} & 20.4 & 0.03 \\
        \hline
    \end{tabular}
\end{table}

\subsection{Semi-supervised}

\myhl{Table \hbox{\ref{tab:semi-supervised_results}} presents MAE and R$^2$ results obtained on the test data with the best-set parameters from Table \hbox{\ref{tab:semi_hyperparameters}}. Results show that $k$-means in the latent space performs better than DEC. Only small values of $k$ were evaluated in \hbox{\cite{xie_deepembedding_2016}} and using a larger $k$ ($>$100) could result in cluster collapse during training (since not all clusters have samples linked to them). The designed features of bag-of-visual words are able to explain some degree of variation in density ($R^2=0.61$), however they not well-suited for estimating income and education, as we discuss later in Section \hbox{\ref{sec:discussion}}. Also in Section \hbox{\ref{sec:discussion}}, we discuss the capabilities of this method and apply explainability techniques.}

\section{Discussion}
\label{sec:discussion}

\subsection{Supervised}

As shown in Table \ref{tab:supervised_results}, resizing the neighborhoods generally produces more accurate results than splitting them into patches. The mean absolute errors in measuring population density and education are better through patching, but their R$^2$s are less than resizing indicating those models do not generalize and explain the variation in density and education as well. A possible explanation for the performance gap in the processing techniques is that resizing the neighborhoods, as opposed to splitting them up, retains more of the spatial and geographic relationships throughout the image amenable to inferring the target metrics. This makes more sense when taken to the logical extreme---a model trained and tested on individual pixels will perform no better than random.
\par
\begin{figure}[ht!]
    \centering
    \includegraphics[width=252pt]{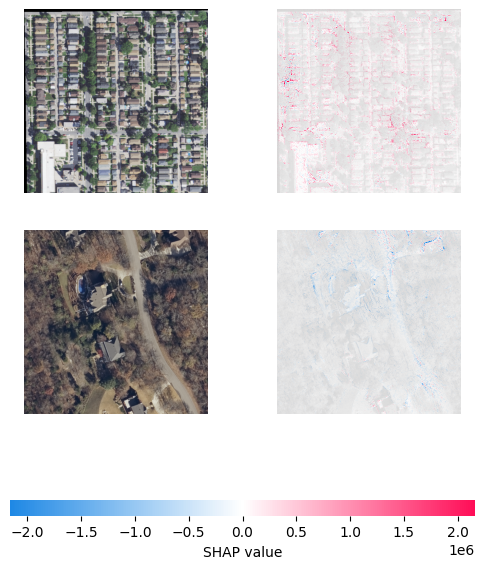}
    \caption{SHAP of two patches from the supervised approach: (top row) within a high-density area of 1814 $\frac{\textrm{ppl}}{\textrm{km}^2}$, and (bottom row) within a low-density area of 11 $\frac{\textrm{ppl}}{\textrm{km}^2}$.}
    \label{fig:shap516-92}
\end{figure}

\myparagraph{CNN interpretation} A frequent criticism of deep learning models is the difficulty of interpreting the relative importance of features in prediction. To help explore what factors contribute to a density model's estimates, we apply SHAP (SHapley Additive exPlanations) saliency map visualizations \cite{lundberg_shap_2017} on patches from two neighborhoods of contrasting density (Fig. \ref{fig:shap516-92}). In Fig \ref{fig:shap516-92}, the first column shows patches in high-density (top) and low-density (bottom) neighborhoods. The second column displays the SHAP values at the pixel level. SHAP values represent the relative importance of features within the selected images. In this example, red pixels represent those that contribute to a higher density estimate while blue pixels contribute to a lower estimate. Fig. \ref{fig:shap516-92} is one of many examples showing man-made structures contributing to a higher density estimate. In the examples, the outlines of smaller dwellings, in particular, are relatively important features. In comparison, in the low-density area, the model demonstrates a tendency to assign a relatively lower density value to larger single-family homes and other features within this more rural area. It is important to note these features would be much less visible from lower-resolution imagery such as Sentinel-2.

\begin{figure}[ht!]
    \centering
    \includegraphics[width=252pt]{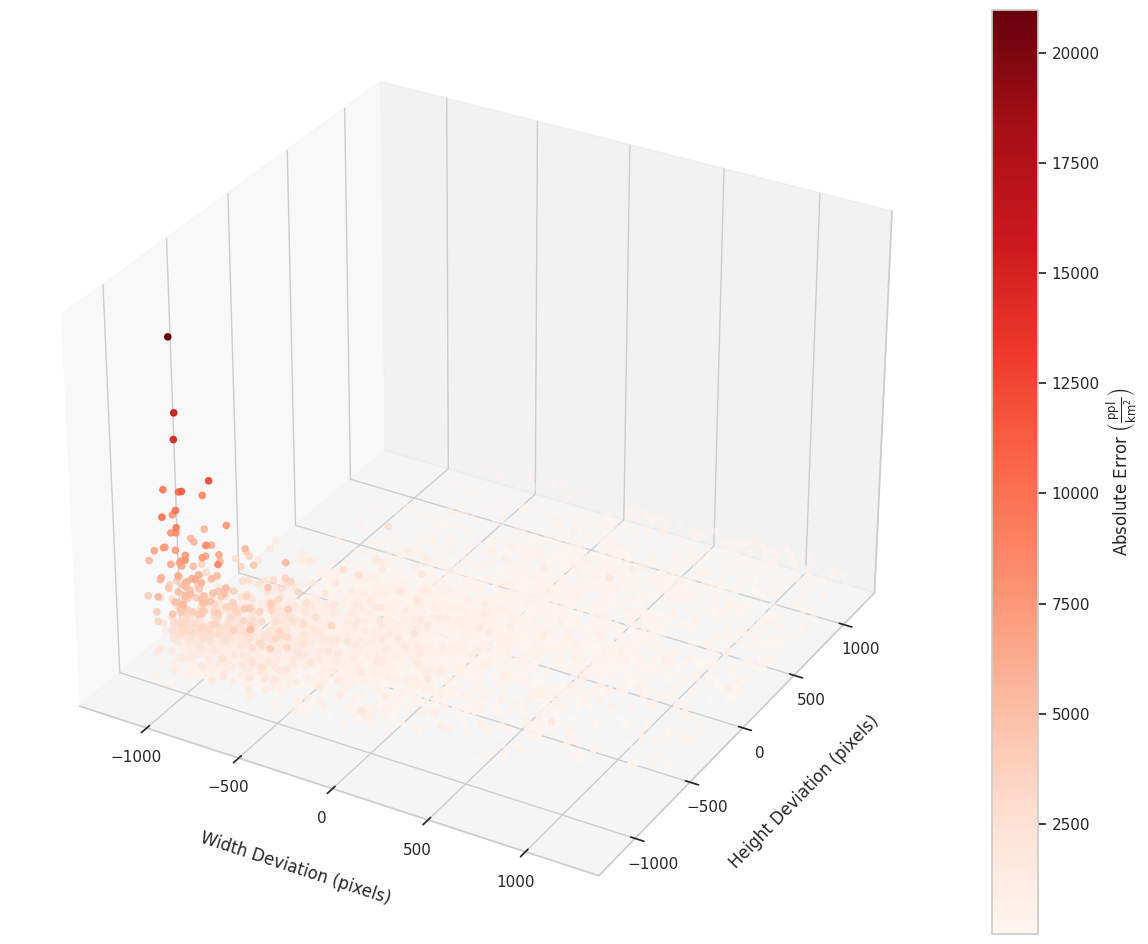}
    \caption{3D plot of the absolute error as a function of width and height deviation from the resized values (when estimating density using the supervised resizing approach).}
    \label{fig:resize_error}
\end{figure}

\myparagraph{Effect of resizing} To better understand the effect of resizing on model estimation, for each image in the density test set, absolute error is plotted against the degree to which the image is resized (Fig. \ref{fig:resize_error}). In Fig. \ref{fig:resize_error}, width (and height) deviation is the difference between an original image's width (and height) and the median value the images are resized to, i.e. 1353 (and 1350 pixels). Interestingly, the result is exponential decay in error as the original images become larger. A possible explanation for this is that larger neighborhoods contain more information, both in the amount of geographic context and in the number of pixels (i.e., in the bilinear interpolation resizing process, a neighborhood image smaller than the median must upsample while an image greater than the median must downsample). This phenomenon also occurs with MHI and education metrics, though to a lesser degree.

\subsection{Semi-supervised}

\myparagraph{Hyperparameters} The semi-supervised method utilizes two important hyperparameters: the dimension of the latent space, $d_Z$, and the number of clusters, $k$. As previously mentioned, different values for the parameters are evaluated, and the ones that provided the best results on the validation data are selected (and displayed in Table~\ref{tab:semi_hyperparameters}). Focusing on the results through $k$-means, it can be seen that the optimal number of clusters is $k =100$ for both MHI and educational attainment, in contrast with the density variable that obtained optimal results with $k = 50$. This is intuitive because more detailed clusters (``visual words'') are necessary to predict the small variations in MHI and education attainment. The latent dimension with the best results is $d_Z = 64$ (the larger of the two tested), and this result indicates that smaller dimensions do not represent the information in the images as accurately.

\begin{figure}
    \centering
    \includegraphics[width=\linewidth]{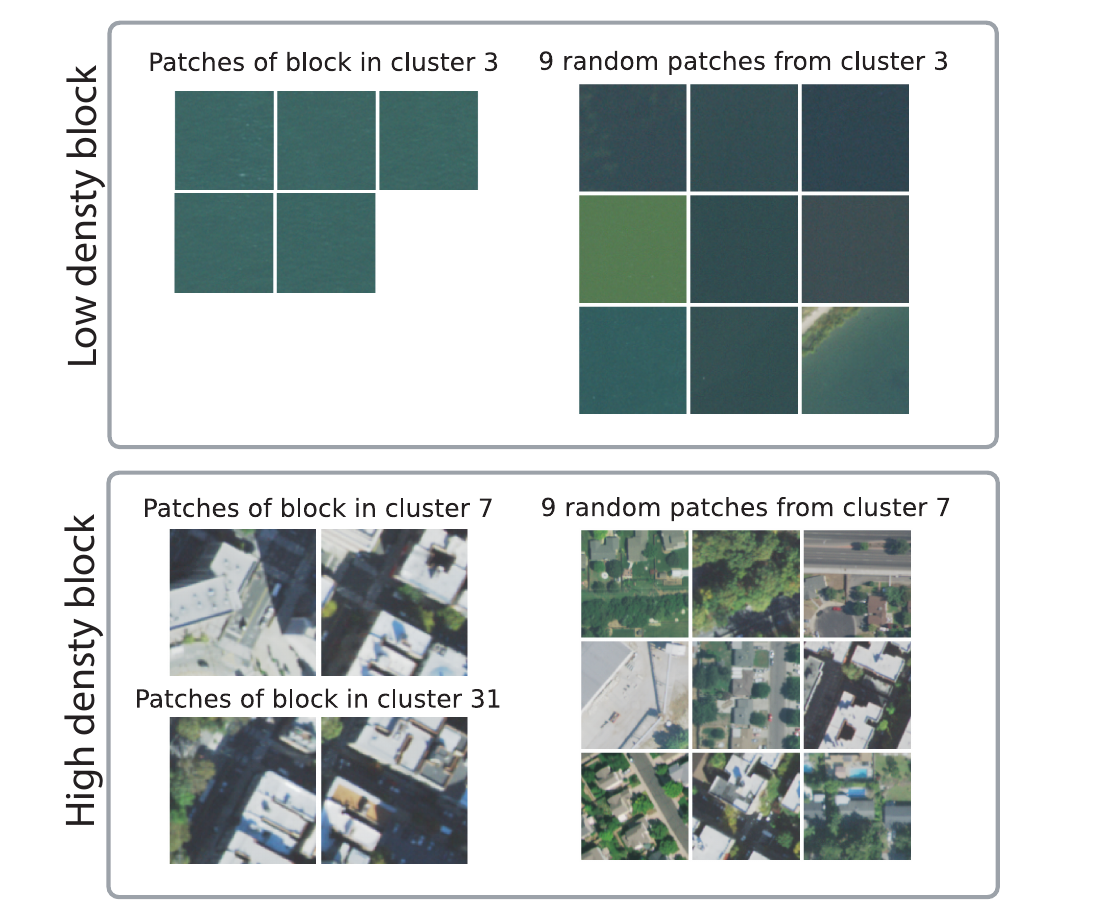}
    \caption{Analysis of the most important features of two neighborhoods in New York using SHAP. The most important cluster for the low-density neighborhood is related to water, and for the high-density neighborhood, two clusters corresponding to densely built areas.}
    \label{fig:interpretation_clustering}
\end{figure}

\myparagraph{Clustering} \myhl{By studying the output clusters, it can be seen that clustering primarily results in groups of undeveloped (natural) geographic areas and groups of urban infrastructure without substantial differentiation within the two groups (though some clusters exhibit congregation of certain feature subtypes such as roads and bodies of water). The method clusters most prominently on natural environment versus built, i.e., the degree of urbanization and development, which are closely correlated with population density but not income or education (at least not within city boundaries in the U.S.). As an illustration, Fig. \hbox{\ref{fig:interpretation_clustering}} shows random samples from two clusters, with cluster 7 containing patches from urban areas but without a particular infrastructure type. Despite its shortcomings, the semi-supervised approach can take advantage of a large unlabeled dataset to create a corpus of bag-of-visual-words features which may be an interesting aspect to exploit in future work.}

\begin{table}[!ht]
    \centering
    \caption{Semi-Supervised Optimal Hyperparameters}
    \label{tab:semi_hyperparameters}
    \renewcommand{\arraystretch}{1.35}
    \begin{tabular}{|c|c|c|c|c|}
        \hline
        \multirow{3}{*}{\textbf{Metric}} & \multicolumn{4}{c|}{\textbf{Method}} \\
        \cline{2-5}
        & \multicolumn{2}{c|}{$k$-means} & \multicolumn{2}{c|}{DEC} \\
        \cline{2-5}
        & $k$ & $d_Z$ & $k$ & $d_Z$ \\
        \hline
        {Density} & 50 & 64  & 100 & 64 \\
        \hline
        {MHI} &  100 & 64  & 20 & 64 \\
        \hline
        {Education} & 100 & 64  & 100 & 32 \\
        \hline
    \end{tabular}
\end{table}

\myparagraph{Interpretation} The proposed semi-supervised methodology presents two learning steps that make use of complex models. First, a deep neural network is employed to cluster, and then a random forest model is used. A t-SNE projection~\cite{maaten_tsne_2008} is used to visualize the latent representation learned by the autoencoder and to comprehend and validate the neighboring relations learned by the network. Similar to the analysis of the supervised method, we use SHAP to study how the random forest regression model interprets cluster features to generate estimations. To exemplify, we select two neighborhoods in New York, one with high density and one with low density. In the low-density neighborhood, the most important features are from cluster 3, and when analyzed in further detail, it is possible to identify that it is a cluster of water patches (shown in Fig.~\ref{fig:interpretation_clustering}). In the high-density neighborhood, the most important features are related to clusters 31, 39, and 7 (despite the neighborhood having no patches in cluster 39). Similarly, by inspecting the patches of these clusters, it can be seen that they are densely built areas with attributes such as high-rise apartments (Fig.~\ref{fig:interpretation_clustering}).

\subsection{Limitations}
For supervised learning, only one resize dimension and patch size are studied, so conclusions can only be drawn based on the arrangements presented. That is, a greater patch size, for example, may lead to better or worse performance than the one chosen (512x512). Our experiments show that despite the semi-supervised approach exhibiting interesting results, it is not able to surpass the performance obtained in supervised learning, particularly when measuring income and education. Clustering is unable to separate nuanced variations in urban infrastructure, which could have been helpful in estimating certain socioeconomic variables such as income and education. Conversely, the supervised models are trained directly on the predicted variables, directly associating image features to metric values. Overall, income and education results from both methods highlight a general limitation of machine learning to extract useful features from aerial imagery correlated with neighborhood-level census variables. Also, it remains a question to what extent the results gathered here in the United States generalize to different social and ecological environments, with or without fine-tuning the models. Finally, it should be noted that many factors contribute to human well-being and not all of them can be quantified, including from \myhl{aerial} or remotely sensed data \cite{worldbank_quant_2019}. To most accurately understand well-being and development, a holistic approach that considers a complex mesh of personal freedoms, institutional capacity and stability, mental and physical health, cultural values, etc. is required \cite{sen_dev_1999}. \myhl{Nevertheless, the techniques employed in this study can serve as a foundation for further refinement, allowing for more precise estimations of socioeconomic variables at a finer granularity than existing literature. This progress can pave the way for future endeavors aimed at estimating the well-being of neighborhoods.}
\section{Conclusion}
\label{sec:conclusion}
Census data require resources and coordination to collect, so are therefore produced relatively infrequently in \myhl{developing} countries. Such data are also usually disseminated with a lag, making it difficult to rapidly assess changes in living standards, especially at local levels. In this work, we explored how well CNNs trained on census data and a semi-supervised clustering approach can estimate census variables in urban neighborhoods throughout the United States. Results show promise in accurately approximating certain metrics (i.e., density), while uncovering limitations for others (i.e., income, education).
\par
Our findings raise several questions for further research including generalizability, whether changes in \myhl{aerial} imagery could be used to forecast changes in \myhl{neighborhood metrics} over time, and how the latest \myhl{aerial} data could be fused with survey-based measures (including ``now-casting'' \cite{engstrom_wellbeing_2022}). Different methods may also be explored at this scale such as the use of semantic models to explicitly extract features such as roads (and their quality), number of buildings (and their type), amount of vegetation, etc. \myhl{(possibly compiled into an ``urban well-being index'')}, for a regressor downstream. As more and more high-resolution \myhl{aerial} imagery products become available (\cite{hand_liftoff_2015,maxar_products}) \myhl{including from cheaply-produced unmanned aerial vehicles (UAVs) deployed outside the U.S. (\hbox{\cite{nezami_drone_2020, retallack_uav_2022}})}, the techniques introduced here provide a foundational benchmark for researchers and reveal potentially fruitful avenues for future work.

\section*{Acknowledgment}
This work was supported by the National Council for Scientific and Technological Development (CNPq, grant \#311144/2022-5), Carlos Chagas Filho Foundation for Research Support of Rio de Janeiro State (FAPERJ, grant \#E-26/201.424/2021), São Paulo Research Foundation (FAPESP, grant \#2021/07012-0), and the School of Applied Mathematics at Fundação Getulio Vargas.



\section*{Code Availability}
Our code is available at \href{https://github.com/VIDA-NYU/GDPFinder}{https://github.com/VIDA-NYU/GDPFinder}.

\bibliographystyle{unsrt}

\bibliography{references.bib}

\begin{thebibliography}{10}

\bibitem{castro_brazil_2023}
DA~Castro and MA~Álvarez.
\newblock Predicting socioeconomic indicators using transfer learning on
  imagery data: an application in brazil.
\newblock {\em GeoJournal}, 88(1):1081--1102, 2023.

\bibitem{un_susdev}
United Nations~Department of~Economic and Social Affairs.
\newblock The 17 goals of sustainable development.
\newblock \url{https://sdgs.un.org/goals}.

\bibitem{psaki_ses_2014}
Stephanie~R. Psaki, Jessica~C. Seidman, Mark Miller, and et~al.
\newblock Measuring socioeconomic status in multicountry studies: results from
  the eight-country mal-ed study.
\newblock {\em Population Health Metrics}, 12(8):8, 2014.

\bibitem{wilson_susdev_2007}
Jeffrey Wilson, Peter Tyedmers, and Ronald Pelot.
\newblock Contrasting and comparing sustainable development indicator metrics.
\newblock {\em Ecological Indicators}, 7(2):299--314, 2007.

\bibitem{sapena_metrics_2020}
Marta Sapena, Luis~A. Ruiz, and Hannes Taubenböck.
\newblock Analyzing links between spatio-temporal metrics of built-up areas and
  socio-economic indicators on a semi-global scale.
\newblock {\em ISPRS International Journal of Geo-Information}, 9(7), 2020.

\bibitem{roland_econdev_2017}
Roland Stephen, Laura Ross, and Nikhil Kalathil.
\newblock Innovative metrics for economic development: Final report.
\newblock {\em Center for Innovation Strategy and Policy. https://eda.
  gov/files/performance/Innovative-Metrics-ED-Report.pdf}, 2017.

\bibitem{worldbank_conflict_2020}
{World Bank Group }.
\newblock Conflicting results: Measuring outcomes in situations of conflict,
  2020.

\bibitem{edmonstonbarry}
Barry Edmonston.
\newblock The case for modernizing the u.s. census.
\newblock {\em Society}, 39(1):42--53, Nov 2001.
\newblock Copyright - Copyright Transaction Publishers Nov/Dec 2001; Last
  updated - 2023-12-04; CODEN - SOCYA6.

\bibitem{SchnakeMahl2020}
Alina~S. Schnake-Mahl, Jaquelyn~L. Jahn, S.V. Subramanian, Mary~C. Waters, and
  Mariana Arcaya.
\newblock Gentrification, neighborhood change, and population health: a
  systematic review.
\newblock {\em Journal of Urban Health}, 97(1):1–25, January 2020.

\bibitem{elvidge_dmspdev_1997}
C.D. Elvidge, K.E. Baugh, E.A. Kihn, H.W. Kroehl, E.R. Davis, and C.W. Davis.
\newblock Relation between satellite observed visible-near infrared emissions,
  population, economic activity and electric power consumption.
\newblock {\em International Journal of Remote Sensing}, 18(6):1373--1379,
  1997.

\bibitem{elvidge_dmspdev_2009}
Christopher~D. Elvidge, Paul~C. Sutton, Tilottama Ghosh, Benjamin~T. Tuttle,
  Kimberly~E. Baugh, Budhendra Bhaduri, and Edward Bright.
\newblock A global poverty map derived from satellite data.
\newblock {\em Computers \& Geosciences}, 35(8):1652--1660, 2009.

\bibitem{burke_susdev_2021}
Marshall Burke, Anne Driscoll, David~B. Lobell, and Stefano Ermon.
\newblock Using satellite imagery to understand and promote sustainable
  development.
\newblock {\em Science}, 371(6535):eabe8628, 2021.

\bibitem{jean_poverty_2016}
Neal Jean, Marshall Burke, Michael Xie, W.~Matthew Davis, David~B. Lobell, and
  Stefano Ermon.
\newblock Combining satellite imagery and machine learning to predict poverty.
\newblock {\em Science}, 353(6301):790--794, 2016.

\bibitem{hu_popindia_2019}
Wenjie Hu, Jay~Harshadbhai Patel, Zoe-Alanah Robert, Paul Novosad, Samuel
  Asher, Zhongyi Tang, Marshall Burke, David Lobell, and Stefano Ermon.
\newblock Mapping missing population in rural india: A deep learning approach
  with satellite imagery.
\newblock In {\em Proceedings of the 2019 AAAI/ACM Conference on AI, Ethics,
  and Society}, AIES '19, page 353–359, New York, NY, USA, 2019. Association
  for Computing Machinery.

\bibitem{yeh_africa_2020}
Christopher Yeh, Anthony Perez, Anne Driscoll, George Azzari, Zhongyi Tang,
  David Lobell, Stefano Ermon, and Marshall Burke.
\newblock Using publicly available satellite imagery and deep learning to
  understand economic well-being in africa.
\newblock {\em Nature Communications}, 11, 05 2020.

\bibitem{goodman_conflict_2021}
Seth Goodman, Ariel BenYishay, and Daniel Runfola.
\newblock A convolutional neural network approach to predict non-permissive
  environments from moderate-resolution imagery.
\newblock {\em Transactions in GIS}, 25(2):674--691, 2021.

\bibitem{runfola_migration_2022}
Daniel Runfola, Heather Baier, Laura Mills, Maeve Naughton-Rockwell, and
  Anthony Stefanidis.
\newblock Deep learning fusion of satellite and social information to estimate
  human migratory flows.
\newblock {\em Transactions in GIS}, 2022.

\bibitem{runfola_testscores_2022}
D.~Runfola, A.~Stefanidis, and H.~Baier.
\newblock Using satellite data and deep learning to estimate educational
  outcomes in data-sparse environments.
\newblock {\em Remote Sensing Letters}, 13(1):87--97, 2022.

\bibitem{stoian_landuse_2019}
Andrei Stoian, Vincent Poulain, Jordi Inglada, Victor Poughon, and Dawa
  Derksen.
\newblock Land cover maps production with high resolution satellite image time
  series and convolutional neural networks: Adaptations and limits for
  operational systems.
\newblock {\em Remote Sensing}, 11(17), 2019.

\bibitem{brewer_roads_2021}
Ethan Brewer, Jason Lin, Peter Kemper, John Hennin, and Dan Runfola.
\newblock Predicting road quality using high resolution satellite imagery: A
  transfer learning approach.
\newblock {\em PLOS ONE}, 16(7):1--18, 07 2021.

\bibitem{vanetten_spacenet_2021}
Adam Van~Etten, Daniel Hogan, Jesus~Martinez Manso, Jacob Shermeyer, Nicholas
  Weir, and Ryan Lewis.
\newblock The multi-temporal urban development spacenet dataset.
\newblock In {\em Proceedings of the IEEE/CVF Conference on Computer Vision and
  Pattern Recognition}, pages 6398--6407, 2021.

\bibitem{helber_eurosat_2019}
Patrick Helber, Benjamin Bischke, Andreas Dengel, and Damian Borth.
\newblock Eurosat: A novel dataset and deep learning benchmark for land use and
  land cover classification.
\newblock {\em IEEE Journal of Selected Topics in Applied Earth Observations
  and Remote Sensing}, 12(7):2217--2226, 2019.

\bibitem{lv_pyshore_2023}
Zhonghui Lv, Karinna Nunez, Ethan Brewer, and Dan Runfola.
\newblock pyshore: A deep learning toolkit for shoreline structure mapping with
  high-resolution orthographic imagery and convolutional neural networks.
\newblock {\em Computers \& Geosciences}, 171:105296, 2023.

\bibitem{asadullah_bangladesh_2012}
Mohammad~Niaz Asadullah and Nazmul Chaudhury.
\newblock Subjective well-being and relative poverty in rural bangladesh.
\newblock {\em Journal of Economic Psychology}, 33(5):940--950, 2012.

\bibitem{helliwell_happy_2022}
J.~F. Helliwell, R.~Layard, J.~D. Sachs, J.-E. De~Neve, L.~B. Aknin, and
  S.~Wang.
\newblock World happiness report.
\newblock Technical report, Sustainable Development Solutions Network,
  \url{https://happiness-report.s3.amazonaws.com/2022/WHR+22.pdf}, 2022.

\bibitem{nces_edattain_2023}
{National Center for Education Statistics}.
\newblock Annual earnings by educational attainment, 2023.

\bibitem{zhan_singlemom_2004}
Min Zhan and Shanta Pandey.
\newblock Economic well-being of single mothers: Work first or postsecondary
  education.
\newblock {\em J. Soc. \& Soc. Welfare}, 31:87, 2004.

\bibitem{jongbloed_euroed_2018}
Janine Jongbloed.
\newblock Higher education for happiness? investigating the impact of education
  on the hedonic and eudaimonic well-being of europeans.
\newblock {\em European Educational Research Journal}, 17(5):733--754, 2018.

\bibitem{zhan_retirement_2002}
Min Zhan and Shanta Pandey.
\newblock {Postsecondary education and the well-being of women in retirement}.
\newblock {\em Social Work Research}, 26(3):171--184, 09 2002.

\bibitem{graetz_africaed_2018}
Nicholas Graetz, Joseph Friedman, Aaron Osgood-Zimmerman, Roy Burstein,
  Molly~H. Biehl, Chloe Shields, Jonathan~F. Mosser, Daniel~C. Casey, Aniruddha
  Deshpande, Lucas Earl, Robert~C. Reiner, Sarah~E. Ray, Nancy Fullman,
  Aubrey~J. Levine, Rebecca~W. Stubbs, Benjamin~K. Mayala, Joshua Longbottom,
  Annie~J. Browne, Samir Bhatt, Daniel~J. Weiss, Peter~W. Gething, Ali~H.
  Mokdad, Stephen~S. Lim, Christopher J.~L. Murray, Emmanuela Gakidou, and
  Simon~I. Hay.
\newblock Mapping local variation in educational attainment across africa.
\newblock {\em Nature}, 555(7694):48--53, March 2018.

\bibitem{greenberg_density_2023}
Michael Greenberg and Dona Schneider.
\newblock Population density: What does it really mean in geographical health
  studies?
\newblock {\em Health \& Place}, 81:103001, 2023.

\bibitem{cramer_density_2004}
Victoria Cramer, Svenn Torgersen, and Einar Kringlen.
\newblock Quality of life in a city: The effect of population density.
\newblock {\em Social Indicators Research}, 69(1):103--116, 10 2004.

\bibitem{mouratidis_oslo_2019}
Kostas Mouratidis.
\newblock Compact city, urban sprawl, and subjective well-being.
\newblock {\em Cities}, 92:261--272, 2019.

\bibitem{xie_poverty_2016}
Michael Xie, Neal Jean, Marshall Burke, David Lobell, and Stefano Ermon.
\newblock Transfer learning from deep features for remote sensing and poverty
  mapping.
\newblock In {\em Proceedings of the Thirtieth AAAI Conference on Artificial
  Intelligence}, AAAI'16, page 3929–3935. AAAI Press, 2016.

\bibitem{zhao_multisourcepov_2019}
Xizhi Zhao, Bailang Yu, Yan Liu, Zuoqi Chen, Qiaoxuan Li, Congxiao Wang, and
  Jianping Wu.
\newblock Estimation of poverty using random forest regression with
  multi-source data: A case study in bangladesh.
\newblock {\em Remote Sensing}, 11(4), 2019.

\bibitem{roy_slums_2020}
Debraj Roy, David Bernal, and Michael Lees.
\newblock An exploratory factor analysis model for slum severity index in
  mexico city.
\newblock {\em Urban Studies}, 57(4):789--805, 2020.

\bibitem{luo_povertymaps_2022}
Eqi Luo, Monika Kuffer, and Jiong Wang.
\newblock Urban poverty maps - from characterising deprivation using
  geo-spatial data to capturing deprivation from space.
\newblock {\em Sustainable Cities and Society}, 84:104033, 2022.

\bibitem{arribasbel_deprivation_2017}
Daniel Arribas-Bel, Jorge~E. Patino, and Juan~C. Duque.
\newblock Remote sensing-based measurement of living environment deprivation:
  Improving classical approaches with machine learning.
\newblock {\em PLOS ONE}, 12(5):1--25, 05 2017.

\bibitem{sandborn_population_2016}
Avery Sandborn and Ryan~N. Engstrom.
\newblock Determining the relationship between census data and spatial features
  derived from high-resolution imagery in accra, ghana.
\newblock {\em IEEE Journal of Selected Topics in Applied Earth Observations
  and Remote Sensing}, 9(5):1970--1977, 2016.

\bibitem{pandey_multitaskpov_2018}
Shailesh Pandey, Tushar Agarwal, and Narayanan~C. Krishnan.
\newblock Multi-task deep learning for predicting poverty from satellite
  images.
\newblock In {\em Proceedings of the AAAI Conference}, 2018.

\bibitem{li_housing_2021}
Guie Li, Zhongliang Cai, Yun Qian, and Fei Chen.
\newblock Identifying urban poverty using high-resolution satellite imagery and
  machine learning approaches: Implications for housing inequality.
\newblock {\em Land}, 10(6), 2021.

\bibitem{engstrom_wellbeing_2022}
Ryan Engstrom, Jonathan Hersh, and David Newhouse.
\newblock Poverty from space: Using high resolution satellite imagery for
  estimating economic well-being.
\newblock {\em World Bank Economic Review}, 36(2):382--412, 2022.

\bibitem{engstrom_density_2020}
Ryan Engstrom, David Newhouse, and Vidhya Soundararajan.
\newblock Estimating small-area population density in sri lanka using surveys
  and geo-spatial data.
\newblock {\em PLOS ONE}, 15(8):1--20, 08 2020.

\bibitem{stevens_popmap_2015}
Forrest~R. Stevens, Andrea~E. Gaughan, Catherine Linard, and Andrew~J. Tatem.
\newblock Disaggregating census data for population mapping using random
  forests with remotely-sensed and ancillary data.
\newblock {\em PLOS ONE}, 10(2):1--22, 02 2015.

\bibitem{anderson_popden_2014}
Weston Anderson, Seth Guikema, Ben Zaitchik, and William Pan.
\newblock Methods for estimating population density in data-limited areas:
  Evaluating regression and tree-based models in peru.
\newblock {\em PLOS ONE}, 9(7):1--15, 07 2014.

\bibitem{deville_mobilepop_2014}
Pierre Deville, Catherine Linard, Samuel Martin, Marius Gilbert, Forrest~R.
  Stevens, Andrea~E. Gaughan, Vincent~D. Blondel, and Andrew~J. Tatem.
\newblock Dynamic population mapping using mobile phone data.
\newblock {\em Proceedings of the National Academy of Sciences},
  111(45):15888--15893, 2014.

\bibitem{sutton_censusheaven_2001}
P.~Sutton, D.~Roberts, C.~Elvidge, and K.~Baugh.
\newblock Census from heaven: An estimate of the global human population using
  night-time satellite imagery.
\newblock {\em International Journal of Remote Sensing}, 22(16):3061--3076,
  2001.

\bibitem{biljecki_popnether_2016}
Filip Biljecki, Ken Arroyo~Ohori, Hugo Ledoux, Ravi Peters, and Jantien Stoter.
\newblock Population estimation using a 3d city model: A multi-scale
  country-wide study in the netherlands.
\newblock {\em PLOS ONE}, 11(6):1--22, 06 2016.

\bibitem{alegana_popmap_2015}
V.~A. Alegana, P.~M. Atkinson, C.~Pezzulo, A.~Sorichetta, D.~Weiss, T.~Bird,
  E.~Erbach-Schoenberg, and A.~J. Tatem.
\newblock Fine resolution mapping of population age-structures for health and
  development applications.
\newblock {\em Journal of The Royal Society Interface}, 12(105):20150073, 2015.

\bibitem{bai_siameseincome_2020}
Ruiqiao Bai, Jacqueline C.~K. Lam, and Victor O.~K. Li.
\newblock Siamese-like convolutional neural network for fine-grained income
  estimation of developed economies.
\newblock {\em IEEE Access}, 8:162533--162547, 2020.

\bibitem{gebru_streetview_2017}
Timnit Gebru, Jonathan Krause, Yilun Wang, Duyun Chen, Jia Deng, Erez~Lieberman
  Aiden, and Li~Fei-Fei.
\newblock Using deep learning and google street view to estimate the
  demographic makeup of neighborhoods across the united states.
\newblock {\em Proceedings of the National Academy of Sciences},
  114(50):13108--13113, 2017.

\bibitem{joachims1998text}
Thorsten Joachims.
\newblock Text categorization with support vector machines: Learning with many
  relevant features.
\newblock In {\em European conference on machine learning}, pages 137--142.
  Springer, 1998.

\bibitem{coates_analysis_2011}
Adam Coates, Andrew Ng, and Honglak Lee.
\newblock An analysis of single-layer networks in unsupervised feature
  learning.
\newblock In Geoffrey Gordon, David Dunson, and Miroslav Dudík, editors, {\em
  Proceedings of the Fourteenth International Conference on Artificial
  Intelligence and Statistics}, volume~15 of {\em Proceedings of Machine
  Learning Research}, pages 215--223, Fort Lauderdale, FL, USA, 11--13 Apr
  2011. PMLR.

\bibitem{bosch_scene_2006}
Anna Bosch, Andrew Zisserman, and Xavier Mu{\~{n}}oz.
\newblock Scene classification via plsa.
\newblock In Ale{\v{s}} Leonardis, Horst Bischof, and Axel Pinz, editors, {\em
  Computer Vision -- ECCV 2006}, pages 517--530, Berlin, Heidelberg, 2006.
  Springer Berlin Heidelberg.

\bibitem{verikas_high_2017}
Samia Bouteldja and Assia Kourgli.
\newblock {High resolution satellite image indexing and retrieval using SURF
  features and bag of visual words}.
\newblock In Antanas Verikas, Petia Radeva, Dmitry~P. Nikolaev, Wei Zhang, and
  Jianhong Zhou, editors, {\em Ninth International Conference on Machine Vision
  (ICMV 2016)}, volume 10341, page 1034120. International Society for Optics
  and Photonics, SPIE, 2017.

\bibitem{kalinicheva_satautoencoder_2020}
Ekaterina Kalinicheva, Jérémie Sublime, and Maria Trocan.
\newblock Unsupervised satellite image time series clustering using
  object-based approaches and 3d convolutional autoencoder.
\newblock {\em Remote Sensing}, 12(11), 2020.

\bibitem{metzler_phenotyping_2023}
A.~Barbara Metzler, Ricky Nathvani, Viktoriia Sharmanska, Wenjia Bai, Emily
  Muller, Simon Moulds, Charles Agyei-Asabere, Dina Adjei-Boadi, Elvis
  Kyere-Gyeabour, Jacob~Doku Tetteh, George Owusu, Samuel Agyei-Mensah, Jill
  Baumgartner, Brian~E. Robinson, Raphael~E. Arku, and Majid Ezzati.
\newblock Phenotyping urban built and natural environments with high-resolution
  satellite images and unsupervised deep learning.
\newblock {\em Science of The Total Environment}, 893:164794, October 2023.

\bibitem{usda_naip}
United States~Department of~Agriculture.
\newblock National agriculture imagery program (naip).
\newblock
  \url{https://catalog.data.gov/dataset/national-agriculture-imagery-program-naip}.

\bibitem{uscensus_acs}
U.S.~Census Bureau.
\newblock American community survey.
\newblock \url{https://www.census.gov/programs-surveys/acs}.

\bibitem{uscensus_5year}
U.S.~Census Bureau.
\newblock When to use 1-year or 5-year estimates.
\newblock
  \url{https://www.census.gov/programs-surveys/acs/guidance/estimates.html}.

\bibitem{uscensus_tiger}
U.S.~Census Bureau.
\newblock Tiger/line® shapefiles.
\newblock \url{https://www.census.gov/cgi-bin/geo/shapefiles/index.php}.

\bibitem{he_resnet_2016}
Kaiming He, Xiangyu Zhang, Shaoqing Ren, and Jian Sun.
\newblock Deep residual learning for image recognition.
\newblock In {\em Proceedings of the IEEE conference on computer vision and
  pattern recognition}, pages 770--778, 2016.

\bibitem{zhou2022comprehensive}
Sheng Zhou, Hongjia Xu, Zhuonan Zheng, Jiawei Chen, Zhao li, Jiajun Bu, Jia Wu,
  Xin Wang, Wenwu Zhu, and Martin Ester.
\newblock A comprehensive survey on deep clustering: Taxonomy, challenges, and
  future directions, 2022.

\bibitem{xie_deepembedding_2016}
Junyuan Xie, Ross Girshick, and Ali Farhadi.
\newblock Unsupervised deep embedding for clustering analysis.
\newblock In {\em Proceedings of the 33rd International Conference on
  International Conference on Machine Learning - Volume 48}, ICML'16, page
  478–487. JMLR.org, 2016.

\bibitem{lundberg_shap_2017}
Scott~M. Lundberg and Su-In Lee.
\newblock A unified approach to interpreting model predictions.
\newblock In {\em Proceedings of the 31st International Conference on Neural
  Information Processing Systems}, NIPS'17, page 4768–4777, Red Hook, NY,
  USA, 2017. Curran Associates Inc.

\bibitem{maaten_tsne_2008}
Laurens van~der Maaten and Geoffrey Hinton.
\newblock Visualizing data using t-sne.
\newblock {\em Journal of Machine Learning Research}, 9(86):2579--2605, 2008.

\bibitem{worldbank_quant_2019}
{World Bank Group}.
\newblock Collateral damage: Pitfalls in quantitative measures of success,
  2019.

\bibitem{sen_dev_1999}
Amartya Sen.
\newblock {\em Development as Freedom}.
\newblock Alfred Knopf, New York, 1999.
\newblock Worldwide publishers: Cappelen Forlag (Norway); Carl Hanser Verlag
  (Germany); China People{\textquoteright}s University Press (China); Companhia
  Das Letras (Brazil); Dost Publishers (Turkey); Editions Odile Jacob (France);
  Editorial Planeta (Spain); Europa Publishers (Hungary); Kastaniotis Editions
  (Greece); Mondadori Editore (Italy); Nihon Keizei Shimbun (Japan); Oxford
  University Press (India); Oxford University Press (UK); Prophet Press
  (Taiwan); Sejong Publishers (Korea); Utigeverij Contact (Holland); Zysk I Ska
  Publishers (Poland) and Dudaj Publishing (Albanian).

\bibitem{hand_liftoff_2015}
Eric Hand.
\newblock Startup liftoff.
\newblock {\em Science}, 348(6231):172--177, 2015.

\bibitem{maxar_products}
Maxar Technologies.
\newblock High-reslution satellite imagery.
\newblock \url{https://www.maxar.com/products/satellite-imagery}.

\bibitem{nezami_drone_2020}
Somayeh Nezami, Ehsan Khoramshahi, Olli Nevalainen, Ilkka Pölönen, and Eija
  Honkavaara.
\newblock Tree species classification of drone hyperspectral and rgb imagery
  with deep learning convolutional neural networks.
\newblock {\em Remote Sensing}, 12(7), 2020.

\bibitem{retallack_uav_2022}
Angus Retallack, Graeme Finlayson, Bertram Ostendorf, and Megan Lewis.
\newblock Using deep learning to detect an indicator arid shrub in
  ultra-high-resolution uav imagery.
\newblock {\em Ecological Indicators}, 145:109698, 2022.

\end{thebibliography}



\end{document}